\documentclass[letterpaper, 10 pt, conference]{style/ieeeconf}  
\IEEEoverridecommandlockouts{}                              

\usepackage[dvipsnames]{xcolor} 
\usepackage[normalem]{ulem} 
\newcommand{\stkout}[1]{\ifmmode\text{\sout{\ensuremath{#1}}}\else\sout{#1}\fi}

\usepackage{graphicx} 
\usepackage{amsmath} 
\usepackage{amssymb}  
\usepackage{physics} 
\usepackage[font=small]{caption} 
\usepackage{subcaption} 
\usepackage{siunitx} 
\usepackage{balance} 

\usepackage{acro}

\usepackage{tikz} 
\usepackage{adjustbox} 
\usetikzlibrary{decorations.pathreplacing, positioning, calc}

\makeatletter
\let\NAT@parse\undefined{}
\makeatother
\usepackage{hyperref} 

\newcommand*{\T}{^\top}

\newcommand*{\SE}[1]{\text{SE} (#1)}

\newcommand*{\q}{\quad}

\newcommand*{\argmin}{\text{argmin}}

\DeclareAcronym{ate}{short=ATE, long=Absolute Trajectory Error}
\DeclareAcronym{rte}{short=RTE, long=Relative Trajectory Error}
\DeclareAcronym{wrte}{short=RTE\(_j\), long=windowed Relative Trajectory Error}
\DeclareAcronym{icp}{short=ICP, long=Iterative Closest Point}
\DeclareAcronym{imu}{short=IMU, long=Inertial Measurement Unit}
\DeclareAcronym{lidar}{short=LiDAR, long=Light Detection and Ranging}
\DeclareAcronym{lio}{short=LIO, long=\ac{lidar}-Inertial Odometry}
\DeclareAcronym{lo}{short=LO, long=\ac{lidar} Odometry}
\DeclareAcronym{slam}{short=SLAM, long=Simultaneous Localization and Mapping}
\DeclareAcronym{atv}{short=ATV, long=All-Terrain Vehicle}
\DeclareAcronym{pca}{short=PCA, long=Principal Component Analysis}

\definecolor{color0}{HTML}{0173b2} 
\definecolor{color1}{HTML}{de8f05} 
\definecolor{color2}{HTML}{029e73} 
\definecolor{color3}{HTML}{d55e00}
\definecolor{color4}{HTML}{cc78bc}
\definecolor{color5}{HTML}{ca9161}
\definecolor{color6}{HTML}{fbafe4}
\definecolor{color7}{HTML}{949494}
\definecolor{color8}{HTML}{ece133}
\definecolor{color9}{HTML}{56b4e9}

\newcommand{\datasetbox}[1]{\raisebox{0.25em}{\fcolorbox{black}{#1}{}}}
\newcommand{\datasetboxinline}[1]{\raisebox{0.35em}{\fcolorbox{black}{#1}{}}}

\newcommand{\figspace}[0]{\vspace{-1.5em}}
\newcommand{\figabove}[0]{\vspace{-0.75em}}

\def\nspace#1{%
  \foreach \index in {1,..., #1} {%
    \vspace{-0.25em}%
  }
}

\newcommand{\edit}[1]{\textcolor{OliveGreen}{#1}}
\newcommand{\remove}[1]{\textcolor{BrickRed}{\stkout{#1}}}

\renewcommand{\edit}[1]{#1}
\renewcommand{\remove}[1]{}

\title{A Comprehensive Evaluation of LiDAR Odometry Techniques}

\author{Easton R. Potokar and Michael Kaess
\thanks{Easton R. Potokar and Michael Kaess are with the Robotics Institute at Carnegie Mellon University, Pittsburgh, PA USA {\tt\footnotesize \{potokar, kaess\}@cmu.edu}}
\thanks{This material is based upon work supported by the U.S. Army Research Office and the U.S. Army Futures Command under Contract No. W911NF20-D-0002. The content of the information does not reflect the position or the policy of the government and no official endorsement should be inferred.}
}

\begin{document}

\maketitle



\begin{abstract}
  \ac{lidar} sensors have become the sensor of choice for many robotic state estimation tasks. Because of this, in recent years there has been significant work done to find the most accurate method to perform state estimation using these sensors. In each of these prior works, an explosion of possible technique combinations has occurred, with each work comparing \ac{lo} ``pipelines'' to prior ``pipelines''. Unfortunately, little work up to this point has performed the significant amount of ablation studies comparing the various building-blocks of a \ac{lo} pipeline. In this work, we summarize the various techniques that go into defining a \ac{lo} pipeline and empirically evaluate these \ac{lo} components on an expansive number of datasets across environments, \ac{lidar} types, and vehicle motions. Finally, we make empirically-backed recommendations for the design of future \ac{lo} pipelines to provide the most accurate and reliable performance.
\end{abstract}


\nspace{3}
\section{Introduction}\label{sec:intro}

Over the last few decades, \acf{lidar} sensors have become the sensor of choice for many robot state estimation and mapping tasks due to their high accuracy, long range capabilities, and complementary nature to other sensors. They have found applications in many platforms from ground vehicles, quadrupeds, and even aerial vehicles, with significant improvement in accuracy and fidelity of the resulting pose estimates and maps.

With this popularity has come a significant increase in \acf{lo} techniques, ranging from methods in feature detection~\cite{zhangLOAMLidarOdometry2014}, initialization~\cite{shanLIOSAMTightlycoupledLidar2020a}, point cloud dewarping~\cite{dellenbachCTICPRealtimeElastic2022}, map aggregation, optimization residual definitions~\cite{segalGeneralizedicp2009}, and many more. With this large number of potential techniques to choose from comes a combinatorial number of potential \ac{lo} ``pipelines'' that can be defined, each with their own strengths and weaknesses.

The current state-of-the-art work usually implements a number of novel techniques along with choosing methods from the before-mentioned list to define a novel \ac{lo} pipeline. This new pipeline is then generally benchmarked against \textit{other pipelines} on a relatively small subset of data to demonstrate an improvement in accuracy, compute efficiency, mapping capabilities, etc. However, there is limited ablation work being done on many of the base choices that underlie these complex state estimation pipelines.

\begin{figure}[t]
  \centering
  \begin{tikzpicture}[
      BOX/.style={rectangle, draw=color0!60, fill=color0!5, very thick, rounded corners=3pt, minimum height=1.5cm, minimum width=2.5cm},
    ]
    \node (origin) {};
    \node[BOX] (dewarp) at (0.0, 2.3) {Dewarp};
    \node[BOX] (feature) [right=of origin, align=center] {Feature\\Extraction};
    \node[BOX] (init) at (0.0, -2.3) {Initialization};
    \node[BOX] (opt) [left=of origin] {Optimization};

    \path[->,very thick]
    (dewarp.east) edge[bend left=30] (feature.north)
    (feature.south) edge[bend left=30] (init.east)
    (init.west) edge[bend left=30] (opt.south);

    \path[->,thick, dotted]
    (opt.north) edge[bend left=30] (dewarp.west)
    (opt.east) edge[bend left=20] ($(init.north)+(-0.25,0.0)$);

    \node (scan) at (-2.75, 2.55) {\acs{lidar}};
    \path[->, very thick] (scan.east) edge ($(dewarp.west)+(0.0,0.25)$);

    \node (imu) {\acs{imu}};
    \path[->,thick, dotted]
    (imu.north) edge (dewarp.south)
    (imu.south) edge (init.north);

    \node (output) at (-2.63, 1.75) {Pose};
    \path[->,very thick] ($(opt.north)+(-0.25,0.0)$) edge (output.south);

  \end{tikzpicture}
  \caption{The basic building blocks that make up a \acf{lo} pipeline. Dotted lines represent optional connections. Few ablation studies have been done comparing these core-level components of \acf{lo} pipelines.}\label{fig:header}
  \figspace{}
  \nspace{1}
\end{figure}
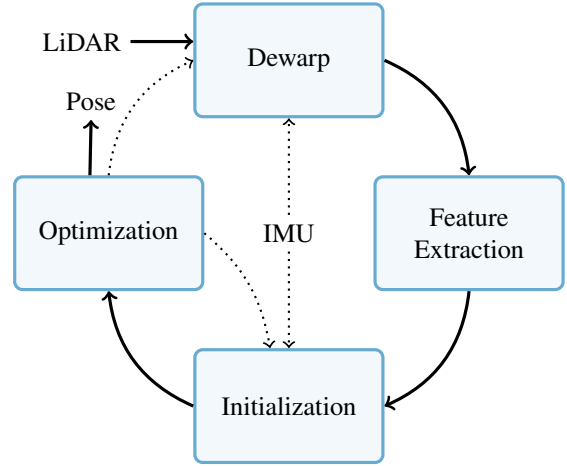

We empirically perform these core-level ablation studies to shed light on what the effects of many of these \ac{lo} components are and provide some recommendations on what techniques should be chosen as a foundation for future \ac{lo} pipelines. Namely, our contributions are:
\begin{itemize}
  \item We provide an introduction to common \ac{lo} techniques to guide a more general audience and provide context and motivation for our later experiments.
  \item We empirically evaluate a number of core \ac{lo} techniques \textit{individually} across a \textit{large} set of trajectories, environments, and \ac{lidar} sensors. This is in contrast to the current prevalent approach of comparing entire \ac{lo} pipelines as a unit on a smaller sample of data.
  \item Based on these results, we supply recommendations on the \ac{lo} components that will provide the best odometry performance for various different scenarios, \ac{lidar} sensors, etc. While empirically-backed and intuitive, some of these results run contrary to many current preferred choices for \ac{lo} pipelines.
\end{itemize}

Finally, we release our python-C++ codebase as open-source\footnote{Available at \href{https://github.com/rpl-cmu/lidar-evaluations/}{https://github.com/rpl-cmu/lidar-evaluations/}} to allow for repeatability and future ablations.

\nspace{1}
\section{Related Works}\label{sec:related_works}
Due to the popularity of \ac{lidar} sensors, numerous \ac{lidar}-based techniques for robot pose estimation have been developed with a combinatorial number of core-level building blocks. There are a number of surveys and reviews that cover these building blocks, but none perform significant ablation studies. Some reviews are analyses of the various pipelines, but they lack any form of empirical results~\cite{huangReviewLiDARbasedSLAM2021,zhang3DLiDARSLAM2024}. Others are focused on general \ac{icp} rather than \ac{lidar}s~\cite{rusinkiewiczEfficientVariantsICP2001} which removes many of the important building blocks such as dewarping and initialization. Finally, some do perform valid empirical experiments, but only compare the \ac{lo} pipelines directly from prior work without any form of ablation studies~\cite{leeLiDAROdometrySurvey2024,jonnavithulaLiDAROdometryMethodologies2021}.
\begin{figure}[t]
    \centering
    \setlength{\fboxsep}{0pt}%
    \setlength{\fboxrule}{0.5pt}%
    \fbox{\includegraphics[width=0.98\columnwidth]{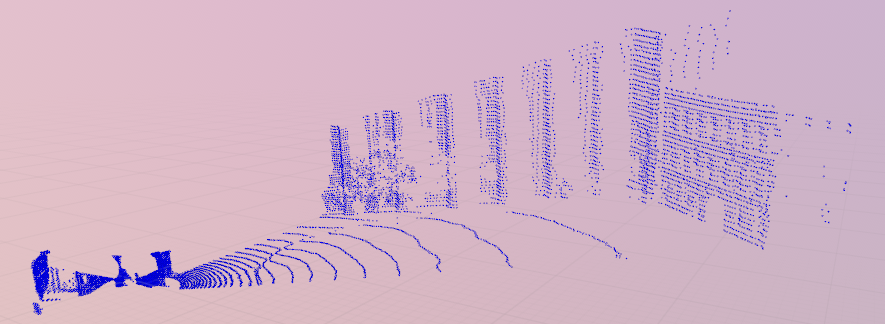}}
    \caption{An example \ac{lidar} scan from the Newer College Stereo-Cam dataset. Note the obvious scanline curves on the ground.}\label{fig:scanline}
    \figspace{}
    \vspace{-0.25em}
\end{figure}

The initial works presenting these \ac{lidar} pipelines also lack clear ablation studies on these components despite the significant amount of diversity in choices such as feature definition and usage, dewarping, initialization, and map aggregation, amongst others.

For example, some methods perform scan matching via vanilla point-to-point \ac{icp}~\cite{vizzoKISSICPDefensePointtoPoint2023}. Other work attempts to model these surfaces as edges~\cite{zhangLOAMLidarOdometry2014} and planar points~\cite{dellenbachCTICPRealtimeElastic2022} to perform feature-based \ac{icp}. Others define additional feature classes using \acs{pca}~\cite{panMULLSVersatileLiDAR2021}. Despite this large diversity in options, and while benchmarks were done comparing pipelines to pipelines, none of these odometry methods or techniques perform comprehensive ablations against the various combinations of potential feature types.

Additionally, there is a large variance in how to define planes and edges in a \ac{lidar} scan. Some use heuristics by summing vectors~\cite{zhangLOAMLidarOdometry2014} along scan lines while others use a more expensive eigenvalue threshold to define planes~\cite{ramezaniWildcatOnlineContinuousTime2022,zhaoSuperOdometryIMUcentric2021}. These are a smaller part of the prior works, but still no experiments investigating the impact of these choices have been done.

Removing motion distortion in a \ac{lidar} scan is also missing empirical comparisons, with common options being a constant velocity model for \ac{lidar}-only methods~\cite{vizzoKISSICPDefensePointtoPoint2023}, \ac{imu} integration for \ac{lio} methods~\cite{shanLIOSAMTightlycoupledLidar2020a}, or using a continuous-time optimization scheme for even higher accuracy~\cite{dellenbachCTICPRealtimeElastic2022,ramezaniWildcatOnlineContinuousTime2022}. Again, while the original authors may have performed ablation studies, little of those experiments have made it into the final published work.

Initialization is also an important part of any \ac{icp} method, but limited studies have been done on its impact. Most \ac{lo} methods use a constant velocity model~\cite{zhangLOAMLidarOdometry2014,vizzoKISSICPDefensePointtoPoint2023} while \ac{lio} methods integrate \ac{imu} integration to generate an initial estimate for \ac{icp}~\cite{shanLIOSAMTightlycoupledLidar2020a,zhaoSuperOdometryIMUcentric2021}. While \ac{imu} integration is assumed to be more accurate, little published work has been done to confirm this, as far as the authors know.

While most of these choices are smaller parts of a larger \ac{lo} pipeline, their joint impact can be significant on the final \ac{lo} pipeline. Due to the lack of a consensus on which methods generally perform the best, we clearly expound on the differences between all these techniques and perform a comprehensive set of empirical studies to shed light on what methods generally result in the highest quality \ac{lo}.

\section{\ac{lidar} Odometry Techniques}\label{sec:methods}
Most \ac{lo} methods rely upon some variant of \ac{icp} for pose estimation between received \ac{lidar} scans. This can be done via scan-to-scan matching or some form of scan-to-map matching, where a map is an aggregate of multiple scans often down-sampled using a voxel grid or kd-tree.

In order to make each execution of \ac{icp} perform independently of previous executions, and to reduce the number of techniques to search over, we focus exclusively on scan-to-scan matching and its related components, and note that the techniques and conclusions found herein should also extend to\edit{, or at least guide development of,} scan-to-map matching.

There is a significant number of options when implementing a \ac{lo} pipeline and even more for a \ac{lio} pipeline, a large subset of which apply to scan-to-scan \ac{icp}. We briefly cover a number of these, both as a precursor to our later experiments and as an introduction for new readers. We also note that while the majority of these techniques are specifically for rotating \ac{lidar}s, a number of conclusions and techniques also apply to solid-state \ac{lidar} sensors as well.

\nspace{1}
\subsection{Dewarping}\label{sec:dewarp}
\nspace{1}
\begin{figure}[t]
    \centering

    \begin{tikzpicture}[
            circle dotted/.style={dash pattern=on .05mm off 2.5mm, line cap=round}
        ]

        \draw [gray!50, very thick] plot [smooth, tension=1] coordinates { (0,0) (0.75,0.75) (2.25,1.1) (3,1.5) };
        \path[->, thick, color=color0] (0,0) edge (3.3,1.3);
        \path[line width=1mm, color=color0, circle dotted] (0,0) edge (3.3,1.3);

        \node at (0,0) {\textbullet};
        \node [anchor=south east] {$X_i$};
        \node at (3,1.5) {\textbullet};
        \node [anchor=south east] at (3,1.5) {$X_{i+1}$};

        \draw [gray!50, very thick, xshift=4.0cm] plot [smooth, tension=1] coordinates { (0,0) (0.75,0.75) (2.25,1.1) (3,1.5) };
        \draw [->, color=color0, thick, xshift=4.0cm] plot [smooth, tension=1] coordinates { (0,0) (0.55,0.55) (2.15,0.8) (3.3,1.3) };
        \draw [color=color0, line width=1mm, circle dotted, xshift=4.0cm] plot [smooth, tension=1] coordinates { (0,0) (0.55,0.55) (2.15,0.8) (3.3,1.3) };

        \node [xshift=4.0cm] at (0,0) {\textbullet};
        \node [anchor=south east, xshift=4.0cm] {$X_i$};
        \node [xshift=4.0cm] at (3,1.5) {\textbullet};
        \node [anchor=south east, xshift=4.0cm] at (3,1.5) {$X_{i+1}$};

    \end{tikzpicture}
    \vspace{-0.5em}
    \caption{Approximating poses between \(X_i\) and \(X_{i+1}\) for dewarping. A constant velocity model is shown on the left, with each dot representing a \(X^i_j\). \ac{imu} dewarping, which more fully captures the motion, is shown on the right.}\label{fig:dewarp_tikz}
    \figspace{}
\end{figure}
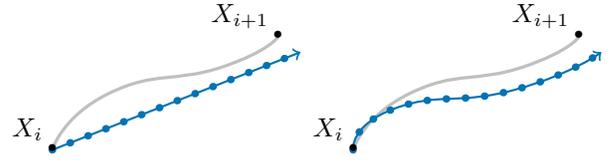

Spinning \ac{lidar} sensors operate by continuously rotating while collecting range measurements along a roughly vertical column. Each element of the vertical column produces a generally horizontal ``scanline'' upon rotation completion, as can be seen in Fig.~\ref{fig:scanline}. At the end of a full rotation, they return all the scanlines, collected into a single ``scan''.

During this rotation, the sensor may undergo motion that can cause a distortion, also called a warp or skew, to be present in the final scans. A common technique is to estimate the motion of the vehicle during the scan and use the motion estimate to ``dewarp'' or ``deskew'' the \ac{lidar} scan.

Assuming a \ac{lidar} scan \remove{began and} was stamped at time \(i\), and a point \(p\) in the scan was taken at time \(j\) and has points represented in the \ac{lidar} frame, the deskewed point \(\tilde{p}\) can be had by using the transformation \(X^i_j \in \SE{3}\) that maps from the \ac{lidar} pose at \(i\) to the \ac{lidar} pose at \(j\),
\begin{align}
    \tilde{p} = X^i_j p.
\end{align}

The following techniques for estimating \(X^i_j\) are shown geometrically in Fig.~\ref{fig:dewarp_tikz}. Estimating \(X^i_j\) is often done by assuming a constant velocity computed using the previous change in pose over change in time and integrating this constant velocity to produce the necessary poses for dewarping.


Another option for estimating \(X^i_j\) is to integrate the high rate \ac{imu} measurements into \(\SE{3}\) poses and interpolate between their timestamps. This requires some form of body velocity and \ac{imu} bias estimation method to perform the \ac{imu} integration, often done in a \ac{lio} pipeline~\cite{zhangLOAMLidarOdometry2014}.

Beyond the scope of this work, another technique is to perform continuous-time pose estimation~\cite{ramezaniWildcatOnlineContinuousTime2022,dellenbachCTICPRealtimeElastic2022}, which then provides the necessary poses continuously for dewarping.

\nspace{1}
\subsection{Initialization}
\nspace{1}
\ac{icp} is known to be extremely sensitive to initialization~\cite{liEvaluationICPAlgorithm2020}. The two most common initialization schemes are constant velocity and \ac{imu} integration.

Constant velocity initialization uses the previous optimized pose delta over the change in time to compute an estimated velocity, and forward propagates this velocity to predict the next pose. If scans come in at a constant rate, this results in using the optimized pose delta at the previous timestep as initialization for the following timestep. \edit{Beyond the scope of this work, more recent work utilizes leasts-squares and a window of past poses to solve for the constant velocity~\cite{ferrari2024mad}}

\ac{imu} integration is generally done in a \ac{lio} pipeline, where body velocity and \ac{imu} biases are also being estimated. Then, using the estimated body velocity and biases, \ac{imu} measurements can be integrated from the previous timestep to the current timestep to provide an initial estimate.
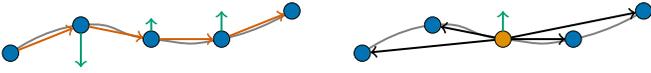
\begin{figure}[t]
    \centering
    \begin{adjustbox}{width=\columnwidth}
        \begin{tikzpicture}[
                mydot/.style={circle, draw, line width=0.05pt, scale=0.7}
            ]
            \draw [gray, thick] plot [smooth, tension=1] coordinates { (-2, -0.2) (-1, 0.2) (0, 0) (1, 0) (2, 0.4) };

            \node[fill=color0][mydot] (x0) at (0, 0.0) {};
            \node[fill=color0][mydot] (x_2) at (-2, -0.2) {};
            \node[fill=color0][mydot] (x_1) at (-1, 0.2) {};
            \node[fill=color0][mydot] (x1) at (1, 0.0) {};
            \node[fill=color0][mydot] (x2) at (2, 0.4) {};

            \path[->, thick, color3] (x_2) edge (x_1);
            \path[->, thick, color3] (x_1) edge (x0);
            \path[->, thick, color3] (x0) edge (x1);
            \path[->, thick, color3] (x1) edge (x2);

            \path[->, thick,color2] (x_1) edge ($(x_1)+(0.0,-0.6)$);
            \path[->, thick,color2] (x0) edge ($(x0)+(0.0,0.3)$);
            \path[->, thick,color2] (x1) edge ($(x1)+(0.0,0.4)$);

            \draw [gray, thick, xshift=5.0cm] plot [smooth, tension=1] coordinates { (-2, -0.2) (-1, 0.2) (0, 0) (1, 0) (2, 0.4) };

            \node[fill=color1, xshift=5.0cm][mydot] (x0) at (0, 0.0) {};
            \node[fill=color0, xshift=5.0cm][mydot] (x_2) at (-2, -0.2) {};
            \node[fill=color0, xshift=5.0cm][mydot] (x_1) at (-1, 0.2) {};
            \node[fill=color0, xshift=5.0cm][mydot] (x1) at (1, 0.0) {};
            \node[fill=color0, xshift=5.0cm][mydot] (x2) at (2, 0.4) {};

            \foreach \x in {x_2, x_1, x1, x2} {
                    \path[->, thick] (x0) edge (\x);
                }

            \path[->, draw=color2, thick, xshift=5.0cm] (x0) edge (0.0, 0.4);

        \end{tikzpicture}
    \end{adjustbox}
    \vspace{-1.1em}
    \caption{Discrete curvature of a 3D parametrized curve shown in gray, of which a \ac{lidar} scanline is a discrete sampling. On the left, the approximated tangents \(\hat{T}_i\) are orange and curvature-scaled normals \(\hat{\kappa}_i \hat{N}_i\) green. On the right, summing the black vectors formed from tangent approximations over a window to estimate \(\hat{\kappa}_i \hat{N}_i\). When \(\lVert\hat{\kappa}_i \hat{N}_i\rVert \) is small, the point is classified as planar and when it is large an edge.}\label{fig:space_curve}
    \figspace{}
    \vspace{-0.25em}
\end{figure}

\nspace{1}
\subsection{Selecting Features}
\nspace{1}
\ac{lidar} features are the method with the highest variance of options used that we will cover. This choice will heavily influence optimization and results, as will be seen in Section~\ref{sec:experiments}. It should be noted, before entering the \ac{lo} pipeline, preprocessing must be done to remove points that are invalid. These include points below the minimum range, above the maximum range, on surfaces parallel to the point ray direction, and when object discontinuity may occur~\cite{zhangLOAMLidarOdometry2014}.

A \ac{lidar} scan is essentially a scan of the 3D surfaces in the environment. Basic \ac{lidar} features attempt to identify the kind of geometry that a point lies on, generally either an edge or a surface. This is usually done by locally estimating some form of the curvature at each point, which defines how curved the geometry is at a given point. Minimally curved geometries correspond to planar points, while maximally curved geometries can correspond to edges. Again, there are a variety of ways to estimate this curvature.

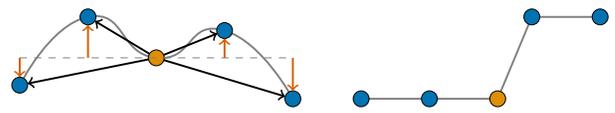
\begin{figure}[t]
    \centering
    \begin{adjustbox}{width=\columnwidth}
        \begin{tikzpicture}[
                mydot/.style={circle, draw, line width=0.05pt, scale=0.7}
            ]
            \draw[white] (-2.5,-0.8) rectangle (7.0,0.8);

            \draw [gray, thick] plot [smooth, tension=1] coordinates { (-2, -0.4) (-1, 0.6) (0, 0) (1, 0.4) (2, -0.6)  };

            \node[fill=color1][mydot] (x0) at (0, 0.0) {};
            \node[fill=color0][mydot] (x_2) at (-2, -0.4) {};
            \node[fill=color0][mydot] (x_1) at (-1, 0.6) {};
            \node[fill=color0][mydot] (x1) at (1, 0.4) {};
            \node[fill=color0][mydot] (x2) at (2, -0.6) {};

            \foreach \x in {x_2, x_1, x1, x2} {
                    \path[->, draw=black, thick] (x0) -> (\x);
                }

            \path[->, color=color3, thick] (-2, 0.0) edge (x_2.north);
            \path[->, color=color3, thick] (-1, 0.0) edge (x_1.south);
            \path[->, color=color3, thick] (1, 0.0) edge (x1.south);
            \path[->, color=color3, thick] (2, 0.0) edge (x2.north);
            \path[color=gray, dashed] (-2, 0) edge (2,0);


            \draw [gray, thick] plot [smooth, tension=0, xshift=5.0cm] coordinates { (-2, -0.6) (-1, -0.6) (0, -0.6) (0.5, 0.6) (1.5, 0.6)  };

            \node[fill=color0, xshift=5.0cm][mydot] (x_1) at (-2.0, -0.6) {};
            \node[fill=color0, xshift=5.0cm][mydot] (x_2) at (-1.0, -0.6) {};
            \node[fill=color1, xshift=5.0cm][mydot] (x0) at (0.0, -0.6) {};
            \node[fill=color0, xshift=5.0cm][mydot] (x1) at (0.5, 0.6) {};
            \node[fill=color0, xshift=5.0cm][mydot] (x2) at (1.5, 0.6) {};

        \end{tikzpicture}
    \end{adjustbox}
    \caption{Various failures of curvature computation. On the left, the vectors horizontally and vertically sum to 0, thus \(\hat{\kappa}=0\) despite it not being a plane. On the right, eigen-based approaches fail to detect the orange point as an edge. Nearest-neighbor eigenvalue will likely miss the far wall as part of the neighborhood, and scanline-based eigenvalue will have two small eigenvalues if the gap is large enough and will detect it as planar. }\label{fig:curvature_diagram}
    \figspace{}
\end{figure}
\subsubsection{Classical Curvature}
One popular technique popularized by LOAM~\cite{zhangLOAMLidarOdometry2014} is to treat each scanline as a discrete sampling from a 3D parametrized curve, also known as a space curve. A simple differencing scheme can then be used to approximate the curvature \(\kappa\in \mathbb{R} \) of the space curve. Given a space curve \(p\) parametrized by arc length \(s\),
\vspace{-0.5em}
\begin{align}
    \begin{split}
        p''(s)             & = T'(s) = \kappa(s) N(s) \\
        \implies \kappa(s) & = \lVert p''(s)\rVert,
    \end{split}
\end{align}
where \(T\) is the \edit{3D} unit tangent vector and \(N\) is the \edit{3D} unit normal. While there is no exact discrete analog to \(\kappa \)~\cite{craneDiscreteDifferentialGeometry2018}, a rough estimate can be done using finite difference methods. With discrete points \(p(i) = p_i \), we can define estimates to the space curve derivatives \(\hat{p}'_i\) and the curvature \(\hat{\kappa}_i\),
\begin{align}
    \begin{split}
        \hat{p}'_i              & = p_i - p_{i-1}                          \\
        \hat{p}''_i             & = \hat{p}'_{i+1} - \hat{p}'_i            \\
        & = p_{i+1} - 2p_i + p_{i-1}               \\
        \implies \hat{\kappa}_i & = \lVert p_{i+1} - 2p_i + p_{i-1} \rVert.
    \end{split}
\end{align}
These vector estimates can be seen on the left of Fig.~\ref{fig:space_curve}. In practice, this is done over a window of points in a scanline to add robustness, which ends up looking like
\begin{align}
    \begin{split}
        \hat{\kappa}_i & = \Big\lVert\frac{1}{n}\sum_{0< j\leq n} (p_{i+j} - 2p_i + p_{i-j}) \Big\rVert \\
        & = \Big\lVert\frac{1}{n}\sum_{-n\leq j \leq n} (p_{i+j} - p_i) \Big\rVert.
    \end{split}
\end{align}
This is visualized in the right of Fig.~\ref{fig:space_curve} where each of the difference vectors can be seen to be summing up to compute \(\kappa_i N_i\). Small values of \(\kappa_i\) then correspond to planar points, while large values correspond to edge points.

There are some assumptions here that \remove{aren't}\edit{are not} necessarily always valid, namely that the curve is parameterized by arc length, and that our discrete curvature necessarily converges to the continuous curvature as the number of points increases. However, this is a well-tested heuristic and empirically has been shown to function well.

\newcommand\crule[3][black]{\textcolor{#1}{\rule{#2}{#3}}}

\begin{table*}[t]
    \centering
    \resizebox{\textwidth}{!}{%
        \begin{tabular}{*{9}{c}}
            \hline
            Dataset                                                         & Color               & Year & Setting        & Device          & Lidar          & Beams & IMU              & Ground Truth               \\
            \hline
            Newer Stereo-Cam~\cite{ramezaniNewerCollegeDataset2020}         & \datasetbox{color0} & 2020 & Outdoor Campus & Handheld        & Ouster OS-1    & 64    & TDK ICM-20948    & Laser Scanner \& ICP       \\
            Newer Multi-Cam~\cite{zhangMultiCameraLiDARInertial2022}        & \datasetbox{color1} & 2022 & Outdoor Campus & Handheld        & Ouster OS-0    & 128   & TDK ICM-20948    & Laser Scanner \& ICP       \\
            Hilti 2022~\cite{zhangHiltiOxfordDatasetMillimeterAccurate2023} & \datasetbox{color2} & 2022 & Indoor Campus  & Handheld        & Hesai PandarXT & 32    & Bosch BMI085     & Laser Scanner \& ICP       \\
            Oxford Spires~\cite{taoOxfordSpiresDataset2024}                 & \datasetbox{color3} & 2024 & Campus         & Backpack        & Hesai QT64     & 64    & Bosch BMI085     & Laser Scanner \& ICP       \\
            Multi-Campus~\cite{nguyenMCDDiverseLargeScale2024}              & \datasetbox{color4} & 2024 & Outdoor Campus & ATV \& Handheld & Ouster OS-1    & 128   & VectorNav VN-200 & Laser Scanner \& Cont-Opt. \\
            HeLiPR~\cite{jungHeLiPRHeterogeneousLiDAR2024}                  & \datasetbox{color5} & 2024 & Urban Road     & Car             & Ouster OS-2    & 128   & Xsens MTi-300    & INS Fusion                 \\
            Botanic Garden~\cite{liuBotanicGardenHighQualityDataset2024}    & \datasetbox{color6} & 2024 & Trail Road     & ATV             & Velodyne VLP16 & 16    & Xsens Mti-680G   & Laser Scanner \& ICP       \\
            \hline
        \end{tabular}
    }
    \caption{Overview of datasets used in evaluation. The majority of sequences from each dataset are included, with the exception of some with motion beyond what is common in a robot trajectory. The color is the color the dataset will use throughout in figures.}\label{table:datasets}
    \figspace{}
    \nspace{1}
\end{table*}
\subsubsection{Scanline Eigenvalue}
An alternative method is to simply compute the covariance of points around a given neighborhood on a scanline,
\begin{align}
    \Sigma_i = \frac{1}{2n} \sum_{-n \leq j \leq n} (p_{i+j} - p_i) (p_{i+j} - p_i)\T.
\end{align}
The eigenvalues of \(\Sigma_i\) will correspond to the main directions the points are spread in. There will always be one small eigenvalue along the scanline, thus we can differentiate planar and edge points with low and high values of the second eigenvalue, respectively.

\subsubsection{Nearest Neighbor Eigenvalue}
This is nearly identical to the prior approach, except rather than using the neighborhood defined by the scanlines, nearest neighbors are found throughout the entire scan. Since there will no longer be a naturally small eigenvalue along the scanline, thresholding can be done using the first eigenvalue, where planar points will have small values once again. Care must be taken to ensure points are on separate scanlines, as points on the same scanline are nearly collinear and will always have a small first eigenvalue.

All of these techniques have various drawbacks. The classical curvature approach can falsely identify a plane if points happened to be symmetrical, as seen in the left of Fig.~\ref{fig:curvature_diagram}. The eigenvalue-based approaches are naturally more computationally heavy and fail to identify edge points due to discontinuities as shown on the right of Fig.~\ref{fig:curvature_diagram}. The nearest neighbor eigenvalue approach should be the most robust due to utilizing points across scanlines to check for planarity, but it does require more parameter tuning across \ac{lidar} beam sizes to ensure neighbors are found on adjacent scanlines and may not function with \ac{lidar}s with fewer beams.

Additionally, some \ac{lo} methods \remove{don't}\edit{do not} bother with features and treat all the points in a \ac{lidar} scan as points. This results in fewer heuristics for computing features and potentially more points to guide the optimization, but it is more sensitive to initialization, as will be seen in Section~\ref{sec:experiments}.

\nspace{1}
\subsection{Optimization}
\nspace{1}
Once features are selected and then matched, the relative pose \(X\in\SE{3}\) that transforms the source scan to the target scan is then optimized for in a nonlinear least squares optimizer with objective,
\vspace{-0.25em}
\begin{align}
    X^\star = \argmin_X \sum_k r_k{(X)}\T \Lambda_k(X) r_k(X)
\end{align}\vspace{-1em}

\noindent where each \(k\) is a pair of matched features, \(\Lambda_k(X)\) is an optional symmetric weighting matrix, and \(r_k\) the residual. \(\Lambda_k\) and \(r_k\) vary based on the type of feature chosen, with definitions as follows. Moving forward, we drop the \(k\) subscript for conciseness.

\subsubsection{Point Features}
Given point \(p_i\) in the target scan and \(p_j\) in the source scan, point-to-point residual is defined as
\nspace{1}
\begin{align}
    r(X)       & = p_i - X p_j \\
    \Lambda(X) & = I.
\end{align}

\subsubsection{Edge Features}
Edge features are optimized by minimizing the portion of the residual that is orthogonal to the estimated edge direction given by \(d_i\). This can easily be done using the annihilator \(A = I - d_i d_i\T \) that removes all vector components that are parallel to \(d_i\). Since \(A\) is a projection matrix \(A = A^2\), we have,
\begin{align}
    r(X)       & = p_i - X p_j        \\
    \Lambda(X) & = A = I - d_i d_i\T.
\end{align}

\subsubsection{Planar Features}
The planar residual has a number of variants~\cite{segalGeneralizedicp2009}. The basic form of point-to-plane can be thought of as the distance of \(p_j\) from the plane given by \(p_i\) and its estimated normal \(n_i\). Similar to edge features, this just requires a custom weighting matrix,
\begin{align}
    r{(X)}\T \Lambda(X) r(X) & = {(n_i \T r)}^2 = (r\T n_i) (n_i \T r) \nonumber \\
                             & = r\T (n_i n_i \T) r                    \nonumber \\
    \implies r(X)            & = p_i - X p_j                                     \\
    \Lambda(X)               & = n_i n_i \T
\end{align}
where \(n_i n_i\T \) is the projection matrix that projects onto \(n_i\). Normal estimation generally occurs by fitting a plane to a subset of points nearby \(p_i\), taking care that points from separate scanlines are used, as points on a single scanline are collinear and induce a degeneracy.

Similarly, a ``pseudo-point-to-plane'' residual can be used to constrain the two orthogonal directions to \(n_i\) as well. The annihilator \(I - n_i n_i\T \) can be used for the projection to these extra dimensions,
\begin{align}
    \begin{split}
        \Lambda(X) & = n_i n_i \T + \epsilon_i  (I - n_i n_i \T) \\
        & = (1- \epsilon) n_i n_i \T + \epsilon I
    \end{split}
\end{align}
for \(\epsilon \in [0, 1]\). Notice that for \(\epsilon = 0\), the standard point-to-plane equation is recovered, while for \(\epsilon = 1\), point-to-point is recovered.

Additionally, a plane-to-plane residual can also be used~\cite{segalGeneralizedicp2009}. It essentially functions as a second residual constraining the target point to source plane given by \(n_j\). Standard plane-to-plane then uses the weighting matrix,
\begin{align}
    \Lambda(X) = n_i n_i \T + R n_j n_j\T R\T
\end{align}
where \(R\) is the rotation in \(X\). A ``pseudo-plane-to-plane'', popularized by Generalized-ICP~\cite{segalGeneralizedicp2009} can then be defined
\begin{align}
    \begin{split}
        \Lambda(X) & = n_i n_i \T + \epsilon  (I - n_i n_i \T)                     \\
        & \q\q + R n_j n_j\T R\T + \epsilon  (I - R n_j n_j\T R\T)      \\
        & = (1- \epsilon) (n_i n_i \T +  R n_j n_j\T R\T) + 2\epsilon I.
    \end{split}
\end{align}

As one can see, there is a combinatorial number of options of feature combinations that can be used, with additional \(\epsilon \) parameters to choose from as well. We explore these, and all other given options, in the following sections.
\begin{figure}[t]
    \centering
    \includegraphics[width=\columnwidth]{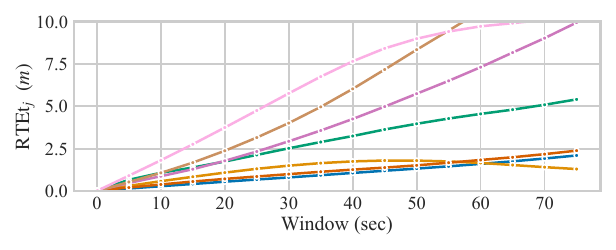}\figabove{}
    \caption{Results of basic scan-to-scan \remove{lidar}\edit{\ac{lidar}} odometry on a number of sequences from across datasets. As the window size increases, \(RTE_j\) generally increases linearly until trajectory effects such as loops begin to have an impact. Dataset colors are as in Table~\ref{table:datasets}.}\label{fig:window}
    \figspace{}
\end{figure}

\section{Experimental Setup}\label{sec:experiments}
\vspace{-0.1em}
We seek to explore the effects of each individual building block independently, and thus attempt to remove as many independent variables as possible in each experiment. Hence, we focus on scan-to-scan matching, noting that the results should transfer to the more accurate scan-to-map matching.

To this end, we also remove the impact of poor body velocity and bias estimation. To do this, we compute bias and body velocity estimates offline for the full trajectory in a factor graph with \ac{imu} preintegration~\cite{forsterOnManifoldPreintegrationRealTime2017} factors and ground truth poses as prior factors. This gives us a consistent and usable bias and velocity estimates for usage in \ac{imu} initialization and dewarping.

Also to remove other independent variables, we use ground truth data to compute velocities any time a constant velocity assumption is used for initialization or dewarping. We also use ground truth initialization in some experiments to remove the impact of poor convergence due to an inaccurate initialization. Further, we evaluate on a large number of datasets and choose a final metric that represents the goal of \ac{lo} well, namely minimizing \remove{long-term} drift.

\nspace{1}
\subsection{Datasets}\label{sec:datasets}
\nspace{1}
Our list of chosen datasets can be seen in Table~\ref{table:datasets}. To ensure that the results generalize well, we prioritized ground truth accuracy as well as maximizing the variety of environments, \ac{lidar} brands and \ac{lidar} beam-size, and motion based on type of vehicle. \edit{All data was loaded using our open-source data-loading library evalio\footnote{Available at \href{https://github.com/contagon/evalio/}{https://github.com/contagon/evalio/}}}

Trajectories range anywhere from around \SI{1}{\minute} to \SI{25}{\minute} long. Due to the quantity of experiments done, \remove{we've}\edit{we have} run exclusively on the first \SI{5}{\minute} of the longer trajectories, or about 3,000 \ac{lidar} scans. We have found this to be a sufficient sample size for drawing conclusions about the various \ac{lidar} approaches we evaluate.

\begin{figure}[t]
    \centering
    \includegraphics[width=\columnwidth]{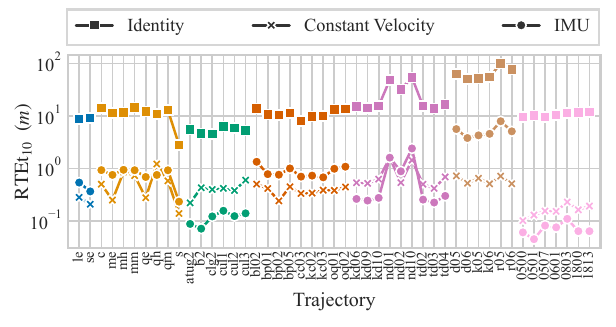}\figabove{}
    \caption{Initialization accuracy with no optimization performed. Both constant velocity and \ac{imu} initialization are computed using ground truth poses and optimized velocities and biases from our offline factor graph. In this idealized scenario, constant velocity is generally more accurate for smooth motion, \ac{imu} initialization for more aggressive motion. Dataset colors are as in Table~\ref{table:datasets}.}\label{fig:init}
    \figspace{}
\end{figure}

\nspace{1}
\subsection{Metric}\label{sec:metric}
\nspace{1}
There are a number of metrics that are commonly used to evaluate odometry methods, the choice of which can have a significant impact on the final results. One common choice is \ac{ate}, which roughly estimates the average global error of the entire trajectory. Given a ground truth pose at timestamp \(i\) \(X_i\) and an estimate given by \(\hat{X}_i\), the translational \ac{ate} is given by,
\nspace{1}
\begin{align}
    \text{ATEt}(X, \hat{X}) = {\left(\frac{1}{N}\sum_{i \in N} {\lVert\text{trans}(X_i \ominus \hat{X}_i)\rVert}^2\right)}^{1/2}.
\end{align}
While very intuitive, \ac{ate} has significant issues. One issue is that a small rotational error at the start of a trajectory will have a significantly larger impact than one at the end of a trajectory~\cite{kummerleMeasuringAccuracySLAM2009}. Further, when dealing with trajectories that revisit locations, a trajectory may have a low \ac{ate} by happenstance if it returns close to the original location regardless of the rest of the trajectory. \edit{This makes \ac{ate} better suited for evaluating results globally and is a poor choice for evaluating drift on local odometry methods.}

A better option for measuring odometry drift is \ac{rte}. Representing a change in pose from timestep \(i\) to \(i+1\) by \(\delta_{i,i+1}\), translational \ac{rte} is given by~\cite{kummerleMeasuringAccuracySLAM2009}
\vspace{-0.9em}
\begin{align}
    \text{RTEt}(\delta, \hat{\delta}) = {\left(\frac{1}{N} \sum_{i = 0}^{N} {\lVert\text{trans}(\delta_{i,i+1} \ominus \hat{\delta}_{i,i+1})\rVert}^2\right)}^{1/2}.
\end{align}
\vspace{-0.1em}
However, most of the datasets that we have used have centimeter level ground truth precision which is near the precision of scan-to-scan odometry, and can result in noisy final metrics.

Instead, since most of the ground truth poses were gathered independently rather than sequentially, we \remove{define}\edit{utilize} \ac{wrte}~\cite{brizi2024vbr}. \ac{wrte} increases the drift error quantities to be larger than the error found in the ground truth, but still avoids the issues that can occur with \ac{ate}. The translational version is defined as
\vspace{-0.1em}
\begin{align}
    \text{RTEt}_{j}(\delta, \hat{\delta}) = {\left(\frac{1}{N-j} \sum_{i=0}^{N-j} {\lVert\text{trans}(\delta_{i,i+j} \ominus \hat{\delta}_{i,i+j})\rVert}^2\right)}^{1/2}
\end{align}\vspace{-0.7em}

\noindent
for some \(j > 0\). Generally, one would expect the value of \ac{wrte} to increase relatively linearly as \(j\) increases. However, if \(j\) is large enough, decreases may appear due to \remove{loops}\edit{tight curves} occurring. An example of this can be seen in Fig.~\ref{fig:window} for the Newer College Multi-Cam~\datasetboxinline{color1} and Botanic Gardens~\datasetboxinline{color6} trajectories, both of which contain a trajectory loop.

We select a window time of \SI{10}{\second}. This ensures that odometric drift is typically above that of the ground truth accuracy, but still avoids large trajectory effects. Additionally, we only show the translational \ac{wrte} moving forward for conciseness since the rotational \ac{wrte} follows similar trends.

As the scale of \ac{wrte} can vary from dataset to dataset due to factors such as environment, \remove{lidar}\edit{\ac{lidar}} beam size, and sensor motion, many figures report the percent change \(C\) in \ac{wrte} to normalize for scale. Thus a change in \ac{wrte} \(a\) over a base case \ac{wrte} \(b\) is given by,
\nspace{1}
\begin{align}
    C = 100 \left(\frac{a - b}{b}\right).
\end{align}%
\nspace{6}
\begin{figure}[t]
    \centering
    \includegraphics[width=\columnwidth]{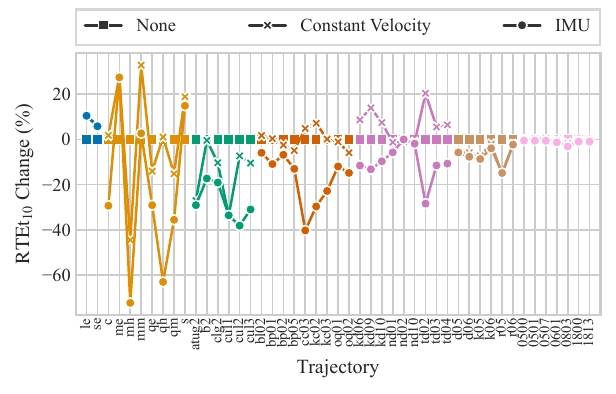}\figabove{}
    \caption{Dewarping experiment showing percent change in \ac{wrte} over no dewarping performed. Constant velocity and \ac{imu} dewarping are computed using ground truth values. Expectedly due to its high sensor rate, \ac{imu} dewarping generally improves performance the most. Dataset colors are as in Table~\ref{table:datasets}.}\label{fig:dewarp}
    \figspace{}
\end{figure}

\section{Experiments}
In the following experiments, we draw conclusions about what feature types should be used, how they should be extracted, what dewarping method provides the best results, and how good of an initialization is needed for convergence to the ground truth optimum. Unless otherwise stated, the base techniques used for each experiment are no dewarping, ground truth initialization, and planar and edge features with the vanilla point-to-plane residual.

\vspace{-0.25em}
\subsection{Initialization}
\nspace{1}
To test initialization accuracy, we compute the \ac{wrte} for each initialization scheme with no optimization, including using the identity, constant velocity, and \ac{imu} integration. For the constant velocity, we use ground truth poses from the prior timestep to compute, and the \ac{imu} integration integrates on top of a ground truth poses with velocity and biases from our offline factor graph optimization. These can be seen as the ideal scenario that will be achieved if the \ac{lo} or \ac{lio} pipeline is functioning properly. Results can be seen in Fig.~\ref{fig:init}.

There are a few trends to note in this figure. First, the motion of the wheeled vehicles in the HeLiPR~\datasetboxinline{color5} and part of the Multi-Campus~\datasetboxinline{color4} datasets are likely notably smoother across timesteps. Due to this, constant velocity is particularly accurate. In contrast, the Newer College Multi-Cam~\datasetboxinline{color1} has some ``medium'' or ``hard'' trajectories that contain aggressive dynamic behavior. In these sequences, constant velocity was naturally worse, with the \ac{imu} integration not being impacted significantly by the aggressive motion.

\begin{figure}[t]
    \centering
    \includegraphics[width=\columnwidth]{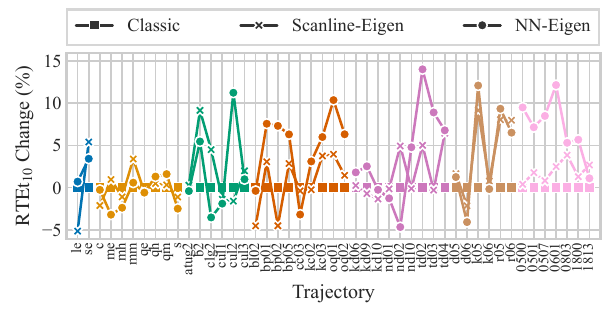}\figabove{}
    \caption{Curvature computation comparison between the classical scanline curvature popularized by LOAM~\cite{zhangLOAMLidarOdometry2014}, scanline-eigenvalue, and nearest-neighbor eigenvalue methods, reported in percent change in \ac{wrte} over classical curvature. All performed very similarly, with the classic method being less computationally expensive. Dataset colors are as in Table~\ref{table:datasets}.}\label{fig:curvature}
    \figspace{}
    \nspace{1}
\end{figure}


While controlled for in this experiment, in real-world deployments it can be argued that there is more that could potentially go wrong with \ac{imu} integration particularly with a low-grade \ac{imu}, namely poor bias or velocity estimation, time-sync, etc. These can impact results, whereas the constant velocity model should remain functional as long as odometry is functioning properly. Thus in cases of smooth motion or a low-quality \ac{imu}, we recommend using constant velocity, and in other scenarios \ac{imu} initialization.

However, as will be seen in later sections, features can be chosen that are less sensitive to initialization and this choice will have a small impact. To reduce the number of changing variables, most of the following experiments are done with ground truth initialization; sensitivity to initialization will follow in its own experiment.

\vspace{-0.25em}
\subsection{Dewarp}
\nspace{1}
We compare dewarping strategies similarly, with ground truth \ac{imu} and constant velocity values, planar and edge features, and ground truth initialization. \edit{Experiments were ran with stamps at the start of scans, which can have a minor impact on constant velocity dewarping.} Results can be seen in Fig.~\ref{fig:dewarp}. We report the percent change in \ac{wrte} between the dewarping methods and no dewarping.

Generally, we found that some form of dewarping was worthwhile, with \ac{imu} dewarping performing the best in the majority of cases. This is unsurprising due to the \ac{imu} measurements capturing higher-rate motion between \ac{lidar} scans. Even the datasets with low-quality \ac{imu}s have a small performance improvement with \ac{imu} dewarping. If \ac{imu} fusion is being performed, \ac{imu} dewarping should definitely be used. If not and computational expense is a constraint, one should considering using no dewarping since the gains for using constant velocity dewarping are more minimal.

\vspace{-0.3em}
\subsection{Curvature}
\nspace{1}
Curvature computations methods, namely classical scan-line based, scanline-eigenvalue, and nearest neighbor eigenvalue methods, are also compared across the trajectory set. We tune the various thresholds for each to emit roughly the same number of features and run experiments with only planar features. The exception is in datasets with small beam sizes, where the nearest-neighbor eigenvalue approach failed to gather as many features due to the large gap across scanlines. Results can be seen in Fig.~\ref{fig:curvature} in percent change from classic curvature.

While they \remove{didn't}\edit{did not} identify identical features, we found that they generally produced very similar results. For those using rotating \ac{lidar}s, we recommend the classical scan-line based approach due to its lower computational overhead and ability to also detect edge features if desired.

\vspace{-0.3em}
\subsection{Pseudo-Features}
\nspace{1}
\begin{figure}[t]
    \centering
    \includegraphics[width=\columnwidth]{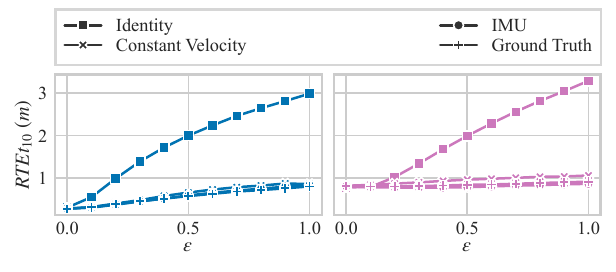}\figabove{}
    \caption{Comparison of \(\epsilon \) parameter in pseudo-point-to-plane optimization across varying initialization strategies. Introduction of any \(\epsilon \) to constrain the orthogonal directions to the normal degrades odometry accuracy. Additionally, point-to-plane optimization is highly insensitive to initialization and the opposite holds for point-to-point optimization. Dataset colors are as in Table~\ref{table:datasets}.}\label{fig:pseudo}
    \figspace{}
\end{figure}

Additionally, we investigate the impact that pseudo-planar features may have on optimization, with results on a pair of trajectories seen in Fig.~\ref{fig:pseudo}. More trajectories were tested, but the majority exhibited similar trends and \remove{aren't}\edit{are not} included for figure clarity. \ac{lo} was run with only planar features and with varying \(\epsilon \) values. The main takeaway from this experiment is that the introduction of constraints on the orthogonal directions to the normal is detrimental to odometry accuracy.

Of note, at \(\epsilon \! \! =\! \!0\) when point-to-plane is recovered, initialization seemed to have a minimal impact, implying a much larger basin of attraction with all initializations converging to the same optimum. At \(\epsilon \!=\!1\) when point-to-point is recovered, the results are highly sensitive to initialization. This is a theme which will be repeated in Section~\ref{exp:features}.

\vspace{-0.3em}
\subsection{Plane-To-Plane}
\begin{figure}[t]
    \centering
    \includegraphics[width=\columnwidth]{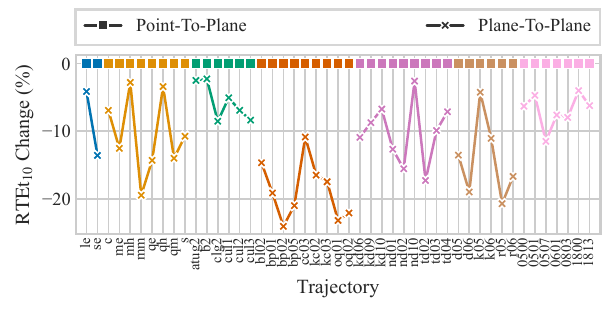}\figabove{}
    \caption{Comparison of point-to-plane and plane-to-plane planar residuals reported in percent change over point-to-plane. Plane-to-plane always outperformed point-to-plane methods. Dataset colors are as in Table~\ref{table:datasets}.}\label{fig:plane_plane}
    \vspace{-0.75em}
\end{figure}

\nspace{1}
We also empirically test plane-to-plane features against the \edit{standard} point-to-plane counterpart. Only planar features are used along with ground truth initialization. We present the results as the percent change over standard point-to-plane in \ac{wrte} in Fig.~\ref{fig:plane_plane}.

This is our result with the most obvious conclusion; when normals can be computed for both target and source point clouds, plane-to-plane residuals will give the best performance, mimicking the results of Generalized-ICP~\cite{segalGeneralizedicp2009}. This intuitively makes sense; it essentially introduces additional constraints and allows for the source cloud normals to help guide the optimization in the weighting matrix \(\Lambda \).

\vspace{-0.3em}
\subsection{Features}\label{exp:features}
\nspace{1}
Finally, we also test the impact that features may have on the final results. We specifically test the combination of point features, planar features, and planar and edge features as these are the most common feature combinations used in prior works. Point features were chosen at the exact locations of edge and planar features to allow for similar quantities to be compared. In practice, there may be significantly more point features. However, this impact is likely to decrease if scan-to-map \ac{icp} is being used and is left for future work.

\begin{figure}[t]
    \nspace{1}
    \centering
    \includegraphics[width=\columnwidth]{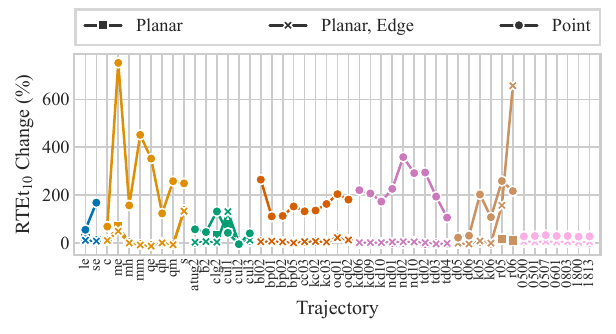}\figabove{}
    \caption{Feature comparison. Results are the percent change from constant velocity to identity initialization. Higher values, like those of the point features, thus demonstrate a high sensitivity to initialization as they improved significantly with better initialization, whereas planar and planar and edge features showed little change. Dataset colors are as in Table~\ref{table:datasets}.}\label{fig:features_point_init}
    \figspace{}
    \vspace{-0.4em}
\end{figure}

The first experiment was run to test the impact of initialization on the various forms of features. Fig.~\ref{fig:features_point_init} shows the relative change from initialization using a constant velocity model to initializing with the identity. Feature types that are sensitive to initialization will result in a increase in \ac{wrte}. It can clearly be seen if something goes poorly during odometry, point features are less likely to recover from a poor initialization than planar features or planar features with edge features. \edit{We also note that point features required more iterations to converge than did planar features, leading to a significant runtime increase.}

\begin{figure}[t]
    \centering
    \includegraphics[width=\columnwidth]{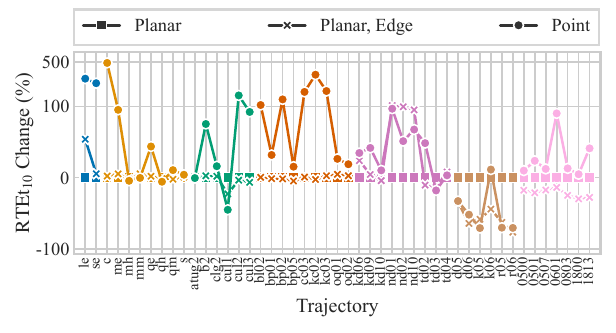}\figabove{}
    \caption{Feature comparison showing relative change to using solely planar features, all initialized with constant velocity. Scale is linear from \(-100\) to \(100\), and log outside that. In most scenarios, edge features had limited positive impact on results, with the exception being in the unstructured HeLiPR~\datasetboxinline{color5} and Botanic Garden~\datasetboxinline{color6} datasets. Dataset colors are as in Table~\ref{table:datasets}.}\label{fig:features_feature_comp}
    \figspace{}
    \nspace{1}
\end{figure}

We additionally test the performance of the various features outright initialized with a constant velocity model. Results are shown in Fig.~\ref{fig:features_feature_comp}. In most scenarios, edge features seemed to improve performance rather minimally and in some cases even degraded performance. The counterexample to this is in the HeLiPR~\datasetboxinline{color5} and Botanic Garden~\datasetboxinline{color6} trajectories, where the environments are extremely unstructured and lack any planar features other than the ground. In these scenarios, edge features detected along the edges of trees helped significantly by adding further constraints.

In any structured environment, we thus recommend using exclusively planar features; in more unstructured environments, using edge \edit{or point} features as well is advisable to aid the optimization.

\nspace{1}
\section{Conclusion}\label{sec:finale}
\nspace{1}
In this work, we have empirically evaluated many of the common core techniques used in \ac{lo} in current state-of-the-art methods across a large sample of data. Specifically, we have covered \ac{icp} initialization, \ac{lidar} scan dewarping, curvature computation methods, various feature optimization residuals, and combinations of common features.

We have shown that planar features with a plane-to-plane residual perform the best in structured environments, with additional features required for unstructured environments. Further, we found little impact in the curvature computation method chosen. Additionally, we showed that \ac{imu} dewarping performs the best, with constant velocity also being worthwhile given there is sufficient compute. We found that constant velocity initialization is excellent if motion is smooth, but \ac{imu} initialization is preferred in cases of aggressive behavior. Finally, we observed that if planar features are used, initialization tends to have a much smaller effect than with edge features.

Future work may include evaluation of further core techniques. These include components such as map aggregation using voxel maps, kd-trees, or others; the optimal way to fuse with \ac{imu} measurements; and which method is most robust for computing planar normals.
\vspace{-0.5em}


\bibliographystyle{style/IEEEtran}
\bibliography{style/IEEEabrv, style/strings-short,ref}

\end{document}